\documentclass[preprint]{elsarticle}

\usepackage{graphicx} % Required for inserting images
\usepackage{booktabs}
\usepackage{appendix}
\usepackage{longtable}
\usepackage{amssymb}
\usepackage{amsmath}
\usepackage{amstext}
\usepackage{float}
\usepackage{dsfont}
\usepackage{array}
\usepackage{wrapfig}
\usepackage{subcaption}

\usepackage{hyperref}

\title{End-to-end Offline Reinforcement Learning for Glycemia Control}
\date{August 2023}

\begin{document}

%\maketitle

\begin{frontmatter}

\affiliation[inst1]{organization={Diabeloop},%Department and Organization
            addressline={17 rue Félix Esclangon}, 
            city={Grenoble},
            postcode={38000}, 
            country={France}}
            
\author[inst]{Tristan Beolet \corref{cor1}}
\cortext[cor1]{Corresponding author: \href{mailto:beolet.tristan@orange.fr}{beolet.tristan@orange.fr}}
\author[inst]{Alice Adenis}
\author[inst]{Erik Huneker}
\author[inst]{Maxime Louis}

\begin{abstract}
    The development of closed-loop systems for glycemia control in type I diabetes relies heavily on simulated patients. Improving the performances and adaptability of these close-loops raises the risk of over-fitting the simulator. This may have dire consequences, especially in unusual cases which were not faithfully -if at all- captured by the simulator. To address this, we propose to use offline RL agents, trained on real patient data, to perform the glycemia control. To further improve the performances, we propose an end-to-end personalization pipeline, which leverages offline-policy evaluation methods to remove altogether the need of a simulator, while still enabling an estimation of clinically relevant metrics for diabetes.
\end{abstract}


\end{frontmatter}

\section{Introduction}

% Ideas:
% personalization is key
% approach fully without simulator
% How to illustrate simulators induce a bias ?

Type I diabetes is an autoimmune disease that leads to the destruction of beta cells in the pancreas. These cells play a crucial role in producing insulin, a hormone responsible for converting blood sugar into energy and regulating blood glucose levels. Without proper treatment, diabetic patients will often experience elevated blood sugar levels, which can lead to both short-term and long-term complications.

Currently, the only course of treatment is insulin therapy: the timely delivery of insulin throughout the day to counteract glucose intake and endogenous production.  In \textit{open-loop} treatment, the insulin delivery is done via boluses--large fast-acting insulin doses--accompanying meals and  basal insulin--a stream of insulin acting steadily through the day--to counteract endogenous glucose production. Closed-loop systems \cite{amadou2021diabeloop, pintaudi2022minimed, cobry2020review, hood2022lived, ekhlaspour2021safety}--or \textit{artificial pancreas}--automatize this procedure: an algorithm connected to both a continuous glucose monitoring device (CGM) \cite{garg2022accuracy, blum2018freestyle} and an insulin pump takes real-time and periodic decisions regarding insulin delivery. This solution not only optimizes glycemia control but also reduces the cognitive load associated with managing the condition.

Given the sensitive nature of these algorithms, whose actions have a direct and potentially harmful impact on the patient's health, the safety and accuracy of the design is of a paramount importance. In vivo testing of such algorithms is also very critical and expensive, hence the design of the algorithm must be made in silico beforehand and with very high confidence. Therefore the primary way to design and validate such algorithms is to use a virtual patient simulator. Such simulators \cite{dalla2007meal} aim at simulating accurately and with enough variety the glycemia patterns of diabetic patients as well as their responses to meals, physical activities, stress, etc. They must also faithfully represent links between patients physiological parameters (e.g. weight, total daily dose etc.) and their glycemia patterns and response to insulin.

Such simulator-based design and validation may be good enough to build and validate high bias/low variance control algorithms, for which the risk is limited. This is how most current commercial closed loop systems have been developed. For instance, using a proportional-integral-derivative (PID) controller, a meal bolus calculator and cutting all insulin delivery in case of upcoming hypoglycemia is a design choice with high bias and low variance which leads to satisfactory performances (see e.g. Table 2 in \cite{emerson2022offline}) in a closed loop, and is fairly robust to simulator biases.

Further improving the quality of a closed loop requires more complex control algorithms. Indeed, perfecting the closed loop control may require to take into account the individual patient physiology, the past patterns observed in the glycemia, the use of additional observables such as sensor data, a learning procedure regarding meals and physical activities, etc. As a consequence, the insulin function of the patient state must take a more flexible form, which is less robust to potential simulator biases. Online reinforcement learning (RL) algorithms \cite{daskalaki2013actor, daskalaki2013personalized, zhu2020basal, zhu2020insulin, fox2019reinforcement, louis2022safe} in particular, which explicitly aim at over-fitting the simulator, can prove particularly risky when used in real life. For instance, on Figure (2a) in \cite{louis2022safe}, one can notice the low basal rate selected by the RL agent even when the glycemia remains above 225 mg/dL during the whole postprandial period. This behavior may be attributed to the simulator on which the model was trained \cite{hovorka2002partitioning}, which is known to never plateau after meal intake, but returns to a steady glycemia state far away from meals. Hence, real life application of this RL agent may lead to lasting hyperglycemia after meals. In \cite{steil2018best}, the authors also show that the description of the hypoglycemia dynamics in the UVA/PADOVA simulator may not be faithful. In the end, relying on a simulator to train online RL algorithms for glycemia control may prove dangerous. Therefore, such algorithms must be used with a lot of additional safety -e.g. far from meals, far from hypoglycemia events or when the patient state/action is sufficiently close to real data \cite{louis2022safe}- which may decrease performances.

To address these shortcomings, offline reinforcement learning \cite{levine2020offline} -RL agents which only use data acquired using some existing policy- is a promising lead. The inability to make counterfactual queries and to explore using online data is replaced with either a policy constraint -e.g. the learned agent must stay close to some existing agent- or an uncertainty-based method -prohibiting optimism in uncertain situations. Such an approach has been used in the context of glycemia control \textit{on simulated data} in \cite{emerson2023offline}. In this paper, the authors made a proof-of-concept: validating the performances and some aspects of safety and robustness of offline RL agents in this context. The fundamental limitation of their work is that it only uses simulated data and validates the RL agents through simulations: they do not demonstrate the applicability of the method on real life data, and do not free themselves from potential simulator limitations.

In this paper, we propose to use offline reinforcement learning to train agents to control the glycemia using \textit{real data} acquired during the commercial exploitation of an existing closed-loop system. Then, we propose an end-to-end pipeline for offline patient-wise personalization of RL agents which \textit{does not require a simulator}. To achieve this, we show how an adaptation of offline policy evaluation (OPE) techniques allows to directly estimate patient-wise clinically relevant metrics such as the time in range (see TIR in Table \ref{tab:metrics}). Results show the validity and efficiency of the approach. To the best of our knowledge, it is the first use of offline RL techniques on real diabetic patient data.

In Section \ref{section:related_work}, we provide an overview of the existing literature on the topic. In Section \ref{section:methods}, we explain how we choose to formalize the problem into an offline RL problem to build population models for control. In particular, we detail the selected offline RL algorithms, the choice of reward function and its importance, the construction of the states and important safety around the agents. In Section \ref{section:experimental_results}, we show the population model results: how they improve over the demonstration data and how they can deal with unannounced meals, in a fully closed loop fashion. Finally, we propose in Section \ref{section:personalization} and end-to-end pipeline for patient-wise offline RL personalization without any use of simulator. 

To summarize, our original contributions are:
\begin{itemize}
    \item The first comparison of offline RL agents for glycemia control using a large set of real data,
    \item An adaptation of FQE to directly estimate key diabetes metrics,
    \item An end-to-end personalization procedure for glycemia control which does not require a simulator -even for validation.
\end{itemize}

\section{Related work}
\label{section:related_work}
%Descriptions of glycemia control online RL papers: can we exhibit some limitations ? Maybe we can at least do it for my paper. Important: if some of the papers do personalization, analyse how it's done (do we do it better ?)

%Descriptions of the only paper on offline RL for glycemia control: give more details than in the intro, and also explain limitations.

%\subsection{Reinforcement Learning for glycemia control}

\paragraph{Indirect RL use} A first group of research studies focuses on optimizing parameters within closed-loop systems through the application of reinforcement learning, resulting in an indirect influence on glycemia control. In \cite{daskalaki2013actor, daskalaki2013personalized}, an actor-critic algorithm is used to refine the insulin to carbohydrate ratio, the meal ratio and the reference basal used by a basal-bolus controller--a simple form of closed loop following the principles of insulin therapy. In \cite{daskalaki2016model}, the authors use an actor-critic agent to tune both the insulin sensitivity and the insulin to carbohydrates ratio. In \cite{zhu2020insulin}, the authors suggest a meal bolus calculator using Deep Deterministic Policy Gradient (DDPG) \cite{lillicrap2015continuous}.

\paragraph{Full RL closed loops} Another group of papers directly trains reinforcement learning agents to calculate optimal insulin deliveries within a closed-loop system. In \cite{zhu2020basal}, deep Q-learning is used to compute the optimal basal value for glycemia control, while meal management is achieved using a traditional meal bolus calculator. Q-learning--using various neural network architectures--is used in \cite{fox2019reinforcement} to train a full closed loop on virtual patients. This baseline work suffers limitations: meal information is not conveyed to the RL agents, there is no safety rule -even in the case of pending hypoglycemia- and the action space is discretized (into several fixed values of basal rates).

\paragraph{RL without a simulator} All of the works cited above use and rely on a virtual patient simulator. While it may be enough for a proof of concept, there remains too much uncertainty to test such agents on real patients -in particular with the management of unusual situations or edge cases. In \cite{emerson2022offline}, the authors investigate the use of offline RL for glycemia control. They use conservative Q-learning \cite{kumar2020conservative}, Twin Delayed DDPG with Behavioral Cloning \cite{fujimoto2021minimalist} and Batch Constrained Deep Q-learning \cite{fujimoto2019off}. To illustrate the potential of offline RL, they train agents on simulated data--collected using a simple proportional-integral-derivative controller--and evaluate the agent on a simulator. In this setting, they achieve promising results: the RL agents--and especially TD3-BC--largely improve over the collection policy. They also show how personalization can be made at the patient level, once again improving over the baseline RL agents. Their work suffers some limitations. First, their offline dataset was collected using a simulator and with additional noise to simulate exploration. This cannot be expected to translate to a real life dataset. Second, the patient state which they used does not contain any individual patient parameter or additional covariable, but merely the glycemia sequence, carbohydrates on board and insulin on board. Third, the UVA/PADOVA simulator \cite{man2014uva} is an overly simplified benchmark for glycemia control \cite{steil2018best}. While useful for demonstration purposes, it does not offer a representative panel of the variety of diabetic situations. In \cite{louis2022safe} for instance, the authors show that near perfect glycemia control can be reached. Finally, their work lack the use of offline policy evaluation methods, which enable to monitor closed loop performances without the use of a simulator.

%\subsection{personalization}

%The second objective of this work is to propose a personalization procedure for RL agents for glycemia control. Such procedures have been tested in the literature. In \cite{mark2022fine, mahmud2022rl}, the authors show a procedure for online fine-tuning of offline RL agents on the d4rl benchmarkcite suited for {fu2020d4rl}: . While interesting, online training - including some exploration - is out of the question for glycemia control, as it may harm the patient. 

\section{Methods}
\label{section:methods}
% \subsection{Problem formalization}

% Reinforcement learning is well-suited for glycemia control:
% \begin{itemize}
%     \item a succession of decision has to be taken: each few minutes the agents needs to choose a basal rate which will be sent to the insulin pump for injection
%     \item the reward is pretty straightforward: the closer the  blood glucose is to 110~mg/dL, the better 
%     \item the reward is delayed: when the basal command is sent, the insulin needs 10 to 15~minutes to enter the blood, and then the all of the insulin is not active straight away    
%     \item the action has to be taken under uncertainty: some hidden factors influencing glycemia control such as illness, sleep, etc. cannot be added in the state
%     \item the action dimension is low: only the basal rate has to be predicted
% \end{itemize}

\subsection{Problem formalization}

%In a closed loop the insulin pump sends a basal rate instruction each time a new blood glucose level value is provided by the CGM to keep the blood glucose level in an acceptable range, in our case each 5 minutes. This interaction mechanism between the patient and the insulin pump can be modeled by a partially observable Markov Decision Process (MDP): at each time step an agent is in a single \textit{state} and takes an \textit{action}, the environment then transitions to a new state and yields a \textit{reward}. In this setting the \textit{agent} is the algorithm controlling the insulin pump, the state is composed of carefully chosen past patterns of glycemia, basal rate and boluses as well as sugar intake and other physiological data ... The action is an insulin basal rate sent to the insulin pump. The reward is a function that evaluates how good a given state is, and is not so trivial to design. Its choice will be explained further in subsection \ref{subsection:reward}. 

%For safety considerations, the maximum basal rate permitted for the agent is 10 U/h. In instances where hypoglycemia is deemed probable, control is handed over to a Hypoglycemia Minimizer (HM). Due to regulatory requirements, the HM is not only incorporated in the training data but also employed in evaluations to determine the efficacy of different RL algorithms in Section \ref{subsection:algo_comparison}, \ref{subsection:baseline_results} and \ref{subsection:unanounced_meals}.

In a closed-loop system, the insulin pump operates with a CGM device -receiving real-time blood glucose level updates at fixed time intervals- and an insulin pump. Its primary objective is to maintain blood glucose within the range 70-180mg/dL. This task is accomplished by computing the appropriate quantity of insulin to deliver each time a new glycemia value is received. This complex decision-making process can be modeled using a Partially Observable Markov Decision Process (POMDP). Therefore, reinforcement learning is a natural idea to build efficient closed loop systems. An \textit{agent} -the algorithm overseeing the insulin pump's control- receives a description of the patient \textit{state} at each time step, selects an \textit{action} and receives the \textit{new state} as well as a \textit{reward}. 

The state should contain any feature which contains information relative to the physiology and glucose level of the patient. Its construction will be detailed further in Section \ref{subsection:state_construction}. While for most classical RL applications, the reward is fixed by the environment itself, we can here design any reward function to optimize. This choice is critical and will be  explained in Section \ref{subsection:reward}.

Meal boluses, ranging from 1 U to more than 15 U, constitute by far the largest instantaneous insulin deliveries. 99.7\% of remaining insulin deliveries found in our data are below 10 U/h. Therefore, to decrease the risk associated with the use of RL, we choose to use a standard meal bolus calculator (as in \cite{amadou2021diabeloop}) to deal with announced meals, leaving the responsibility of all remaining insulin to the RL agent. We model each RL agent action as an insulin rate between 0 and 10 U/h. This way, the potential harm that may be caused by the RL agent is much lower than if the agent was allowed to prescribe meal boluses. To allow for some flexibility, we use a reasonably safe meal bolus calculator, which is unlikely to overestimate meal boluses. Doing so, the RL agents can still increase postprandial insulin, indirectly adjusting the meal bolus.

Additionally, since hypoglycemia events are the greatest immediate threats to the patients, in any situation where an impending risk of hypoglycemia is detected, all insulin deliveries are stopped. The criterion to measure the risk of hypoglycemia is a linear regression on the glycemia signal, and a condition on the predicted glycemia 15min-1h in the future.

Meal boluses and hypoglycemia prevention is used in all the experiments made in section \ref{section:experimental_results}.

%The agent action corresponds to the basal insulin rate sent to the insulin pump, a choice from the authors. Then a bolus is just a high basal rate lasting one time step. In addition a meal bolus calculator -similar to the one used with the DBLG1 system- is used to help manage meal periods. As a consequence, the basal rate predicted by the agent stays low: $<$10U/h. This approach as the advantage to be safe:  the agent cannot send to much insulin to fast. In situations where an impending risk of hypoglycemia is detected by linear regression, the basal injection is stopped. In all the experiments made in section \ref{subsection:algo_comparison} and \ref{subsection:baseline_results}, meal boluses and hypoglycemia prevention are used. 

\subsection{RL algorithms}

%TO DO: explain DQN, maybe SAC and TD3 ? 
In Reinforcement Learning (RL), the primary objective for an agent is to maximize the cumulative return $G$ defined as the sum of the discounted rewards $G_t = \sum_{k=0}^{\infty} \gamma^kR_{t+k+1} $, where $R_t$ is the reward the agent receives at time step $t$.

%$\gamma$ is the discount factor which controls the effective horizon of the agent: $\frac{1}{1-\gamma}$ and represents how much the agent takes into account future rewards when making decisions. With $\gamma = 0.99$ the effective horizon is 3 hours in our case which is a suitable time window for glycemia control.
The discount factor $\gamma$ determines how much the agent values future rewards compared to immediate rewards, controlling the effective horizon. In our context, with $\gamma = 0.99$, the agent time horizon is of order $\frac{1}{1-\gamma}\simeq 8h$ since each time step lasts 5 minutes. This time horizon is relevant because close to the maximum duration of the effects of meals and boluses on the glycemia.

%Instead of estimating $G$ directly, the state-action function or Q-function $Q(s,a)  = \mathbb{E}\big[G_t \vert s, a\big]$ is used to estimate how much an action $a$ will yield return in state $s$. 

The state-action function, denoted as the Q-function, defined as $Q(s,a) = \mathbb{E}\big[G_t \vert s, a\big]$ is an analog of G. This function provides an estimate of the expected return when taking action $a$ in state $s$.

%Glycemia control is a complex task because it's influences by a multitude of factors such as activity, illness, sleep, stress, medication, diet and the unique patient physiology. How those factors impact glucose consumption is not fully understood yet % Need some refs here. 

%Acieving optimal glycemia control is a difficult challenge due to its susceptibility to numerous influencing factors, including physical activity, illness, sleep patterns, stress levels, medication, dietary choices, and the unique physiological characteristics of each patient. Therefore building an accurate internal model of the patient may prove impossible. Model free RL makes no assumptions about the environment the agent interacts with and thus is a valid candidate for glycemia control. 

While most RL applications to glycemia control are online and use a simulator, we argue that offline reinforcement learning offers much better perspectives. First, as discussed in the introduction, online approaches for glycemia control may lead to over-fitting of the biases of the simulator. Second, the now widely used closed loops systems have enabled the acquisition of vast quantities of data. Third, the existing closed loop systems already give solid results on a wide variety of patients, and it is clear that all closed loop systems should follow policies which are reasonably close to existing systems. Indeed, existing systems perform well, are safe and follow a rationale from usual insulino-therapy, broadly validated by diabetologists. For all these reasons, offline RL algorithms constitute excellent approaches to improve on existing closed loop systems.

Capturing individualized glucose-insulin dynamics through model-based approaches is challenging due to inherent physiological variability. Therefore, model-based RL methods may be challenging. On the other hand, model-free methods offer a computationally efficient alternative, eliminating the need to predict these intricate dynamics at each decision step. Furthermore, the empirical performances of deep learning-backed model-free techniques, as evidenced by successes with DQN, A3C \cite{mnih2016asynchronous}, and TRPO \cite{schulman2015trust} in diverse domains, underscore their potential for this application. Coupled with the safety imperative of an offline approach—given the risks of online biases and catastrophic outcomes from incorrect dosing—model-free offline RL emerges as a robust solution. It leverages historical patient data, offering policies that are data-driven and individually adaptive without the pitfalls of model-induced errors. We focus on model-free offline RL methods in the rest of the paper.

\subsection{Offline RL algorithms}

Offline RL offers the advantage of learning from large real-world datasets but the impossibility to explore raises new challenges. A particular concern arising from this is distribution shift: as the newly trained policy is different from the behavior policy -policy used to collect the data-, the actions taken are also different from those the behavior policy would take. Different actions lead to different next states which can lead to a different distribution of encountered states -from the one observed from the behavior policy- and can lead to overconfidence in some states, posing potential risks.

Each offline algorithm used in this paper mitigates distributional shift differently:
\begin{itemize}
    \item \textbf{Batch Constrained Q-learning (BCQ)}: a variational auto-encoder generates counterfactual queries, which represent a set of actions following the distribution of the training data, as described in \cite{fujimoto2019off}. The selected action is determined by choosing the one with the highest Q value, and clipped double Q learning is employed to limit Q-value overestimation.
    \item \textbf{Conservative Q-learning (CQL)}: Instead of learning an approximation of the Q-function, CQL estimate a lower bound function of the Q-function to reduce overestimation \cite{kumar2020conservative}. In-distribution state-action pair are assigned higher q values than out of distribution ones. 
    \item \textbf{Twin Delayed DDPG with Behavioral Cloning (TD3-BC)}: An extension of the actor critic algorithm TD3 \cite{fujimoto2018addressing}. A behavioral cloning term in TD3 actor loss is used to select actions that are often seen in the training data \cite{fujimoto2021minimalist}. Clipped double Q-learning and Q-function smoothing are also used. 
    %\item \textbf{TD7}: Adaptation of the TD3-BC algorithm using behavioral cloning to prevent overestimation and introduces embeddings of state and state action pair \cite{fujimoto2023sale}, and a Loss Adjusted Prioritized (LAP) \cite{fujimoto2020equivalence} replay buffer.
\end{itemize}

\subsection{State construction}
\label{subsection:state_construction}

% Constraints: 
% - Have to be in real data
% - Have to be in the simulator

% Found the optimal by hyperparameter tuning, multiple iterations

Given the data at hand, the available features to construct the state are described in Table \ref{table:state_vars}.

\begin{table}[h!]
\caption{Data used to build the agent state.}
\centering
\renewcommand{\arraystretch}{1.5}
\begin{tabular}{p{3.5cm} p{5.5cm} p{2cm}}
\toprule
\textbf{Variables} & \textbf{Description} \\
\midrule

Glycemia History & Past records of blood sugar levels\\
Insulin History & Past records of insulin injection rates\\
Insulin Metrics &  IOB (insulin on board, computed as in openAPS, see \cite{OpenAPS2023}, and TDD (total daily dose, representing daily insulin needs)\\
Carbohydrate Metrics & COB (carbohydrate on board, indicating undigested sugar intake) \\
Time Metrics & Current time of day\\
Physiological Metric & Body weight\\
\bottomrule
\end{tabular}
\label{table:state_vars}
\end{table}

Glycemia history offers valuable insights into past blood sugar levels, enabling the RL agent to identify trends and patterns. This historical context is instrumental in predicting future glucose levels and assessing the success of prior insulin interventions. Insulin history complements glycemia history by providing a comprehensive record of insulin injection rates over time. Understanding how the patient has responded to insulin dosages in the past is useful for adapting recommendations to their evolving insulin sensitivity and individual insulin requirements. The IOB quantifies the insulin that has already been subcutaneously injected but has not yet had an action in the blood stream, influencing future glycemic responses. The TDD, on the other hand, offers an overview of daily insulin intake, aiding the agent in tailoring recommendations for glycemic control throughout the day. The COB introduces the crucial concept of undigested carbohydrate intake. When patients consume carbohydrates, blood sugar levels do not immediately reflect this intake. COB accounts for this delayed effect, allowing the agent to anticipate and mitigate potential glycemic spikes resulting from recent carbohydrate consumption. The consideration of time metrics, specifically the time of day, is essential. Sensitivity to insulin and liver activity naturally fluctuate throughout the day due to circadian rhythms and mealtime variations \cite{grant2022multi, belsare2023understanding}. Finally, the body weight can impact insulin sensitivity and metabolism.

The process of state construction involves careful optimization to determine the essential co-variables and the appropriate time horizon for time series data. After meticulous analysis, our research has revealed that the most effective state configuration comprises IOB, COB, TDD, time of day, and a one-hour window for the time series.

%To construct an optimal state, an optimization has been conducted to determine the co-variables that must be kept and the time horizon for the time series. In the end, the most effective state is composed of IOB, COB, TDD, time of day and the time series contain one hour of previous data

All features are normalized using zero-mean unit-variance or min-max scaling.

%\begin{itemize}
%    \item Glycemia: $X \leftarrow \frac{X - 40}{360}$
%    \item Time: $X \leftarrow \frac{X}{24\times60}$
%    \item Insulin: $X \leftarrow \frac{X}{60.0}$
%%    \item IOB: $X \leftarrow \frac{X}{25.0} $
%    \item COB:$X\leftarrow \frac{X}{100.0}$
%\end{itemize}

% It's important to note that the basal rate predicted by the agent as the action as been clamped to the interval [0,10] insulin unit per hour. 

\subsection{Metrics}
\label{subsection:method_algo_comparison}
% TO DO: reporter les citations de l'article sur le offline RL et le controle de glycemie
%In our performance assessment, we utilize clinical metrics widely acknowledged in evaluating glycemia control algorithms. These metrics, detailed in Table \ref{tab:metrics}, ensure that our agents' glycemia control aligns with regulatory and clinical standards. TIR is the primary objective the agent is trying to maximize.

%These metrics link our machine learning agent to clinical contexts, tying data-driven decisions to real-world outcomes. By assessing our agents against clinical benchmarks, we ensure our algorithms meet strict clinical and regulatory standards for glycemia control. This underscores the clinical validity of our solutions and a possible future deployment in a commercial device.

In our performance assessment, we employ clinical metrics, as detailed in Table \ref{tab:metrics}, universally recognized for evaluating glycemia control algorithms. The Time In Range (TIR) is the primary metric for closed loops. These metrics bridge machine learning agent to real-world outcomes, ensuring trained agents offer clinically satisfying glycemia control. The clinical targets for each metrics can be seen in the consensus found in \cite{de2022ispad}.

\begin{table}[H]
\caption{Blood Glucose Metrics}
\centering
\begin{tabular}{|p{2.3cm}|p{7cm}|p{1.5cm}|}
\hline
\textbf{Metric} & \textbf{Description} & \textbf{Target} \\
\hline
Time-in-Range (TIR) & Percentage of time where the glycemia is in the range: 70-180 mg/dL. Increased TIR is strongly associated with a reduced risk of developing micro-vascular complications \cite{beck2019validation}. & $>70\%$ \cite{battelino2019clinical} \\
\hline
Time-Below-Range (TBR) & Percentage of time where the glycemia is lower than 70 mg/dL also called time in hypoglycemia. & $<4\%$ \cite{battelino2019clinical},\cite{holt2021management}  \\
\hline
Critical Time-Below-Range (TBR$<$54) & Percentage of time where the glycemia is lower than 54 mg/dL. & $<1\%$ \cite{american20216}, \cite{holt2021management} \\
\hline
Time-Above-Range (TAR) & Percentage of time where the glycemia is greater than 180 mg/dL, also called time in hyper glycemia. & $<25\%$ \cite{battelino2019clinical} \cite{holt2021management} \\
\hline
Coefficient of Variation (CV) & Relative dispersion of blood glucose values around their mean. A high value of the coefficient of variation entails a higher probability of vascular tissue damage \cite{ceriello2019glycaemic}. & $<36\%$ \cite{danne2017international}, \cite{holt2021management} \\
\hline
Mean glycemia & Mean blood glucose value (mg/dL). A high value increases the probability of dementia \cite{crane2013glucose} and cardio-vascular tissue damage \cite{ceriello2019glycaemic}  & As low as possible \\
\hline
\end{tabular}
\label{tab:metrics}
\end{table}

\subsection{Reward function design}
\label{subsection:reward}

Unlike common RL tasks, there is no predefined reward function for glycemia control. The choice of reward function is absolutely crucial, as it drives entirely the behavior of the RL agents. For simplicity, we use reward functions which only depend on the current glycemia value. More sophisticated functions, depending on some history of glycemia or on the current rate of insulin may be used e.g. to penalize glycemia variations or too high insulin rates.

\begin{figure}%{r}{0.6\textwidth}
%\begin{subfigure}[b]{0.5\textwidth}
    \centering
    \includegraphics[width=0.5\linewidth]{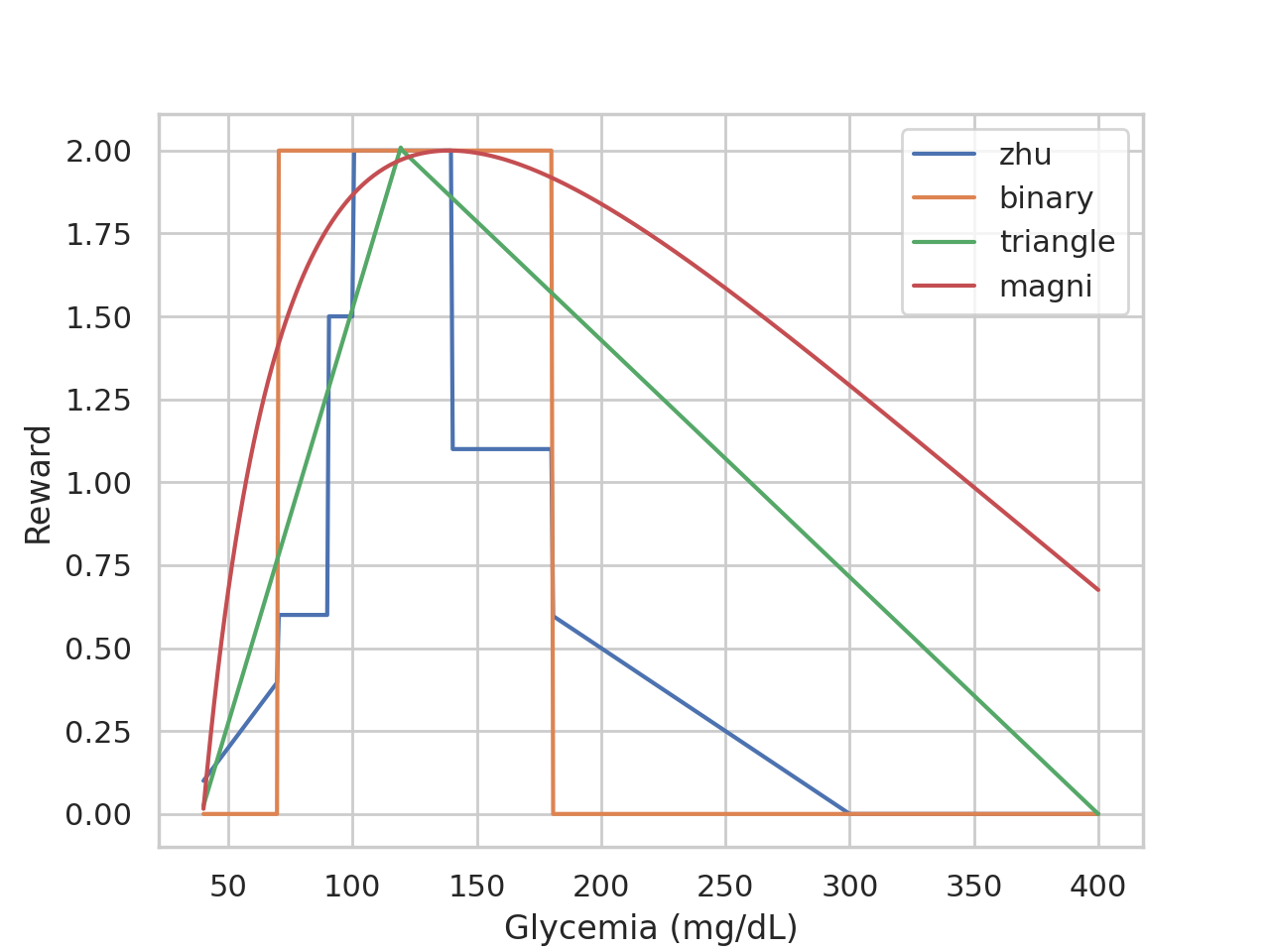}
    \caption{Candidate reward functions.}
    \label{fig:reward_functions}
  %\end{subfigure}%
  %\begin{subfigure}[b]{0.5\textwidth}
   % \centering
   % \includegraphics[width=\linewidth]{images/rewards/custom_piece_wise_function.png}
   % \caption{Ours}
   % \label{fig:custom_reward}
  %\end{subfigure}
  %\caption{Comparison of different reward functions }
  %\label{fig:reward_functions_vs_metrics}
\end{figure}

We show on Figure \ref{fig:reward_functions} a list of reward functions analyzed in this paper. The simplest is the binary reward whose maximization amounts to TIR maximization. Rewards from \cite{zhu2020basal, emerson2022offline, amadou2021diabeloop} are other alternatives. These rewards represent different trade-offs between hypoglycemia and hyperglycemia situations, as well as different tolerances to mild deviations from the ideal glycemia of 110 mg/dL.

\begin{figure}[H]
  \centering
  \includegraphics[width=\linewidth]{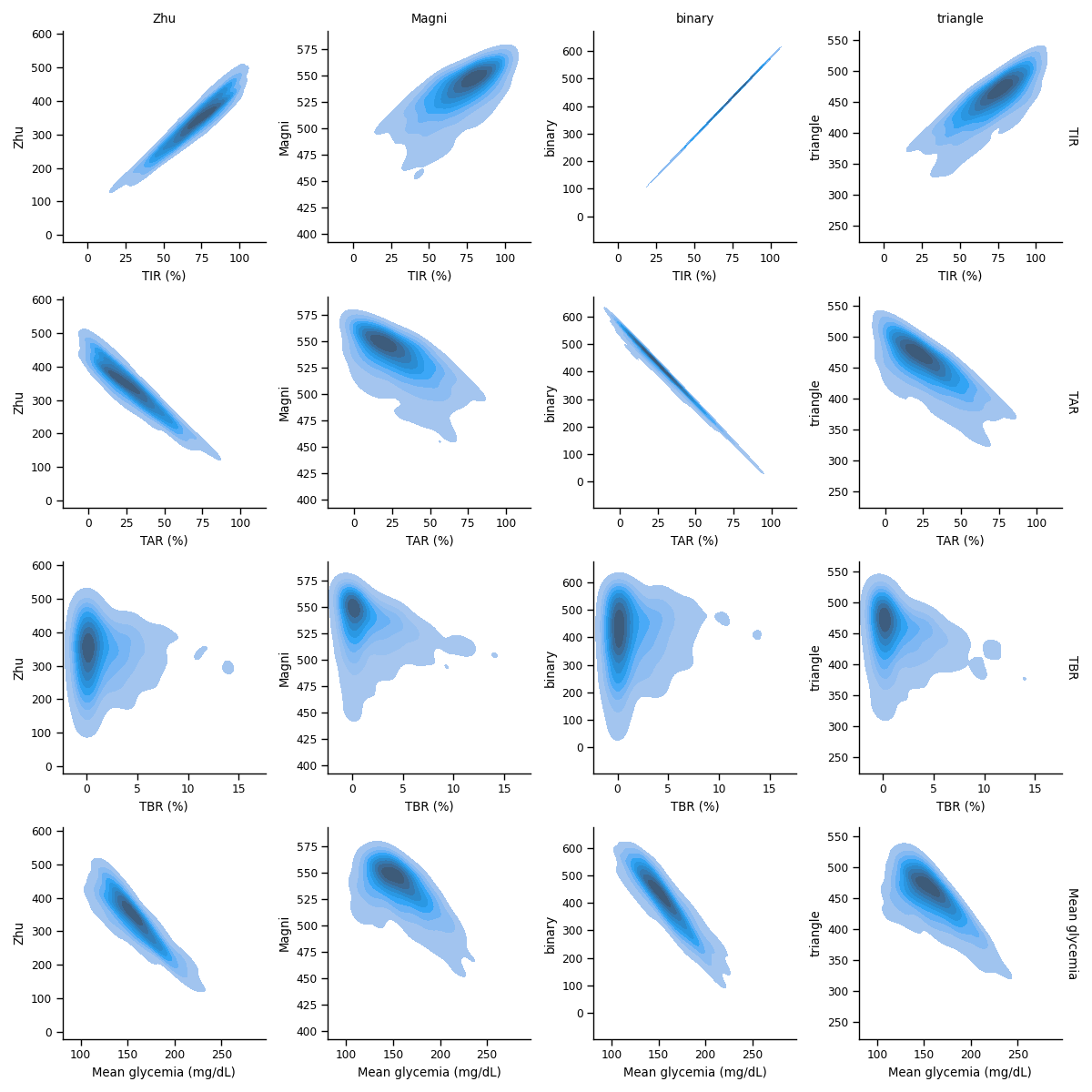}
  \caption{Correlation between clinical metrics for glycemia control against different reward functions, high correlation means that increasing the training reward will increase / decrease the clinical metric as wanted.}
  \label{fig:reward_functions_vs_metrics}
\end{figure}

To determine which reward function is the most appropriate without unnecessary computations, we look at how the performance metrics described in the previous section vary for each patient-day with respect to the sum of rewards on the same days. Such an analysis is shown on Figure \ref{fig:reward_functions_vs_metrics}. Magni and triangle rewards are the ones that correlate the least with TIR and TBR. Empirical testing showed that binary reward was not informative enough for high blood glucose values sometimes resulting in the inability to reduce the blood glucose level after meals.

%TO CORRECT: Even though Magni reward function achieved acceptable results, the TBR ratio was way to high, that's why we introduced a new reward function empirically designed to achieve a TBR lower than 4\%. Among all tested reward functions, agents trained with our reward function have been found to achieve the best balance between hypoglycemia and hyperglycemia.

%The rewards function that seems to have a good shapes are: triangle reward, binary\_reward and reward\_1. 
%Training with those metrics lead to too much hypoglycemia events (time below range should stay below 4\%), and thus a new piece wise reward function have been introduced. This shape is easily editable compared to reward\_Magni, and enabled to reach an acceptable time below range. 
%Binary reward was not informative enough 

%TO DO: write this 

\section{Experimental results: population models}
\label{section:experimental_results}

We provide the implementation of all offline RL agents used and some of the training/evaluation pipelines \hyperlink{https://github.com/TristanBEOLET/offline-RL}{on this repository}.

\subsection{Data acquisition and pre-processing}
\label{subsection:data}

The data used to train and evaluate the RL agents has been collected through real life commercial usage of the DBLG1 artificial pancreas \cite{amadou2021diabeloop}. This closed loop system equips more than 10000 patients, with some patients wearing the system for more than 2 years. Among the ones agreeing to have their data collected, we selected 100 at random to form a training set for the RL agents. We filter to keep only days during which the closed loop was activated for more than 70 \% of the time. Table \ref{table:training_data_info} contains general characteristics regarding the data used. Overall, there are 6.9 million transitions in the dataset.

Note that the behavior policy is not \textit{exactly} the DBLG1 algorithm as:
\begin{itemize}
    \item Some open loop points may remain within the data (e.g. when disconnection with the pump or the CGM occurs)
    \item Even when the closed loop is activated, the patient has the possibility to modify boluses suggested by the algorithm (and does so in 1.4 \% of cases) or to manually prescribe boluses (that represents 1.4 \% of total boluses).
    \item Some parameters of the DBLG1 algorithm can be adjusted by the patients so as to improve the (perceived) quality of the closed loop. It may or may not improve the TIR/TAR/TBR metrics, depending on the patient particular sensitivity.
\end{itemize}
We argue that it is an advantage when using the data to train offline RL agents. Indeed, the state/action distribution in the data is larger than with the raw behavior policy, bringing more information during training of offline RL agents. An ablation study, removing the occurrences of manual and modified boluses for instance, would allow to quantify this effect, though beyond the scope of the work presented here.

\begin{table}[h!]
    \caption{Training data details.}
    \centering

    \begin{tabular}{|c|c|}
        \hline
                                                      &        Mean ± Std \\
        \hline
                                                  Age &       44.8 ± 13.0 \\
                                          Weight (kg) &       77.1 ± 17.5 \\
                                 Total daily dose (U) &       44.0 ± 18.1 \\
          Proportion of manually modified boluses (\%) &         1.4 ± 1.8 \\
        Proportion of manually prescribed boluses (\%) &         1.4 ± 2.4 \\
                                    TIR (\%) &        68.4 ± 9.9 \\
                                 TBR (\%) &         1.3 ± 1.6 \\
                            TAR (\%) &       30.3 ± 10.1 \\
                             CV &         33.0 ± 4.5 \\
                             Mean glycemia (mg/dL) &      160.8 ± 14.6 \\
                              Number of observed days &     284.1 ± 161.2 \\
                               Number of observations & 69 000 ± 39 000 \\
        \hline
    \end{tabular}
    \label{table:training_data_info}
\end{table}

\subsection{Algorithm comparison}
\label{subsection:algo_comparison}

We perform an hyper-optimization of TD3-BC, BCQ and CQL algorithms on the data presented in \ref{subsection:data}. Working with a limited compute budget -for economical and ecological reasons- we iteratively optimize several features in an A/B testing manner: the state composition (length of glycemia and insulin history), the reward function used and several hyper-parameters critical to each algorithm such as the RL/BC trade-off of TD3-BC. 

Each trained agent performances are evaluated online on the simulator. While we argue that in silico validation should not be the gold standard of closed loop validations and design, it is still very informative, and we view it as a necessary -but not \textit{sufficient}- condition for algorithm validation. In any case, we will show in the next section how to leverage offline policy evaluation methods to measure closed-loop candidate performances. 

As mentioned before:
\begin{itemize}
    \item When a meal is declared, the RL action is overridden by a meal bolus computed using the calculator in \cite{amadou2021diabeloop}.
    \item When there an impeding risk of hypoglycemia is detected, the RL agent is deactivated and no can be sent to the patient. 
\end{itemize}

The best performance metrics for each algorithm are shown on Table \ref{table:rl_comparison}. TD3-BC and BCQ outperform the behavior policy in terms of TIR, TAR and mean glycemia. TD3-BC is better than the behavior policy for all metrics but the coefficient of variation, which remains within acceptable bounds. This shows the ability of offline RL algorithms to improve over the behavior policy, confirming the findings of \cite{emerson2022offline}.

\begin{table}[ht]
\caption{\centering Comparison of the offline algorithm for glycemia control, values are Mean $\pm$ Std. Best value for each metric is in bold}
\centering
%\tiny
\begin{tabular}{llll}
\toprule
{} &             TIR &            TBR &         TBR$<$54 \\
\midrule
BCQ   &   70.4$\pm$7.59 &  3.87$\pm$4.17 &  0.98$\pm$1.85 \\
CQL   &  57.8$\pm$10.52 &  9.93$\pm$9.57 &  5.13$\pm$6.71 \\
TD3-BC &   \textbf{74.38$\pm$7.3 }&   \textbf{2.73$\pm$3.71} &  \textbf{0.86$\pm$ 1.84} \\
Behavior policy   &  69.89$\pm$ 7.97 &  3.59 $\pm$ 3.65 &  1.44 $\pm$ 2.37
 \\
\midrule
{} &              TAR &               CV &     Mean Glycemia \\
\midrule
BCQ   &   25.72$\pm$7.01 &   39.38$\pm$7.84 &  \textbf{147.94$\pm$11.79} \\
CQL   &  32.27$\pm$11.76 &  42.94$\pm$11.92 &   150.75$\pm$20.5 \\
TD3-BC &   \textbf{22.89$\pm$5.86} &   36.90$\pm$ 7.79 &   148.62$\pm$10.82 \\
Behavior policy   &  26.51 $\pm$7.04 &  \textbf{33.55 $\pm$ 7.13} &  156.59 $\pm$ 9.28 \\
\bottomrule
\end{tabular}
\label{table:rl_comparison}
\end{table}

\begin{figure}[H]
  \centering
  \includegraphics[width=\textwidth]{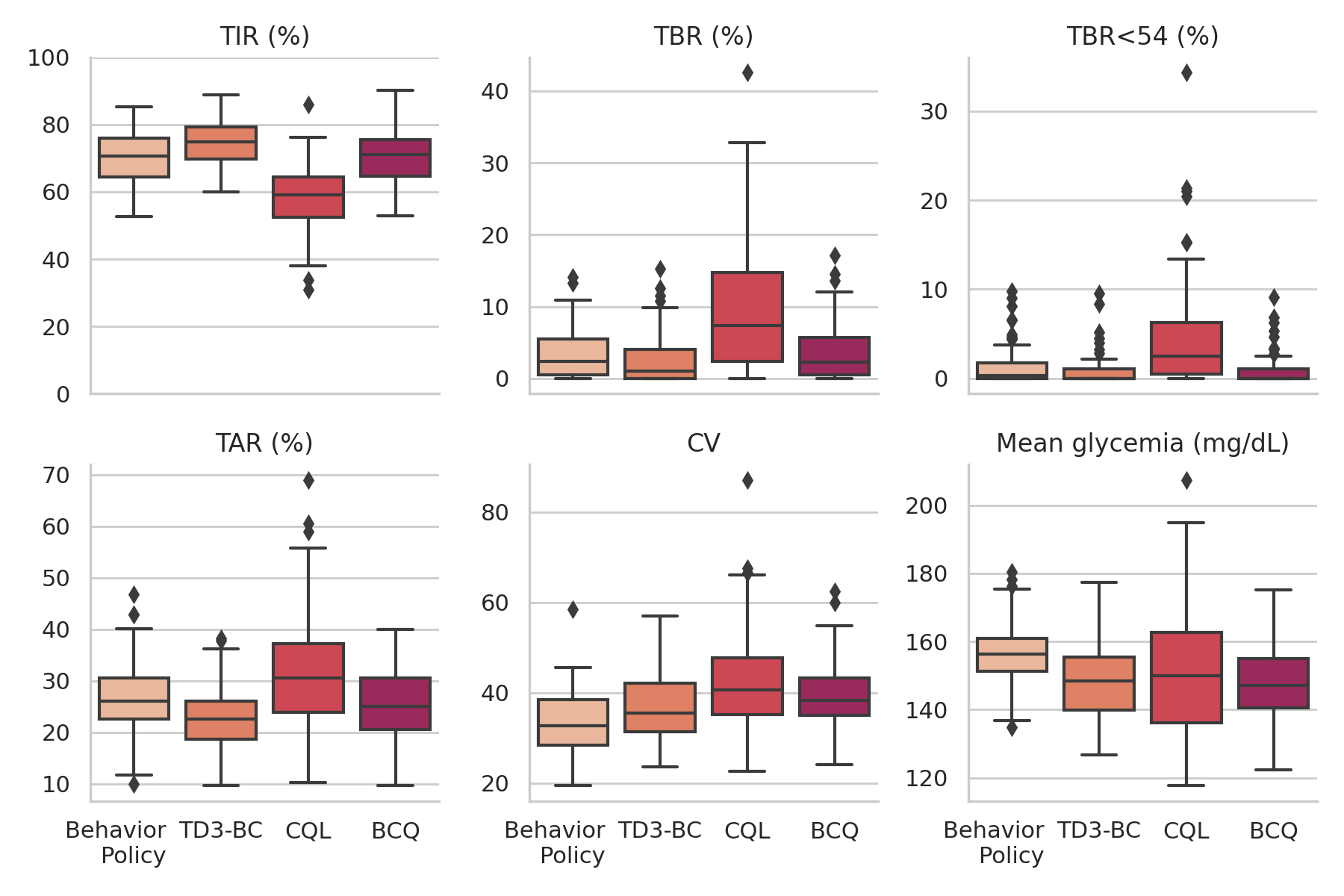}
  \caption{Comparison of the different policies on the simulator.}
  \label{fig:comparison_algos}
\end{figure}

\subsection{Population model analysis}
\label{subsection:baseline_results}

The best-performing model -which we refer to as population model from now on-, TD3-BC, found in the comparison above has been compared on the simulator with the behavior policy.  We can see from Table \ref{table:rl_comparison} that the TD3-BC agent is able to simultaneously improve the TIR by 4.49\%, and the TBR by  0.86\%. Note that in any case, the inter-patient performance variability remains quite high e.g. TIR range from around 50\% to above 80\%.

\begin{figure}[H]
  \centering
  \includegraphics[width=\textwidth]{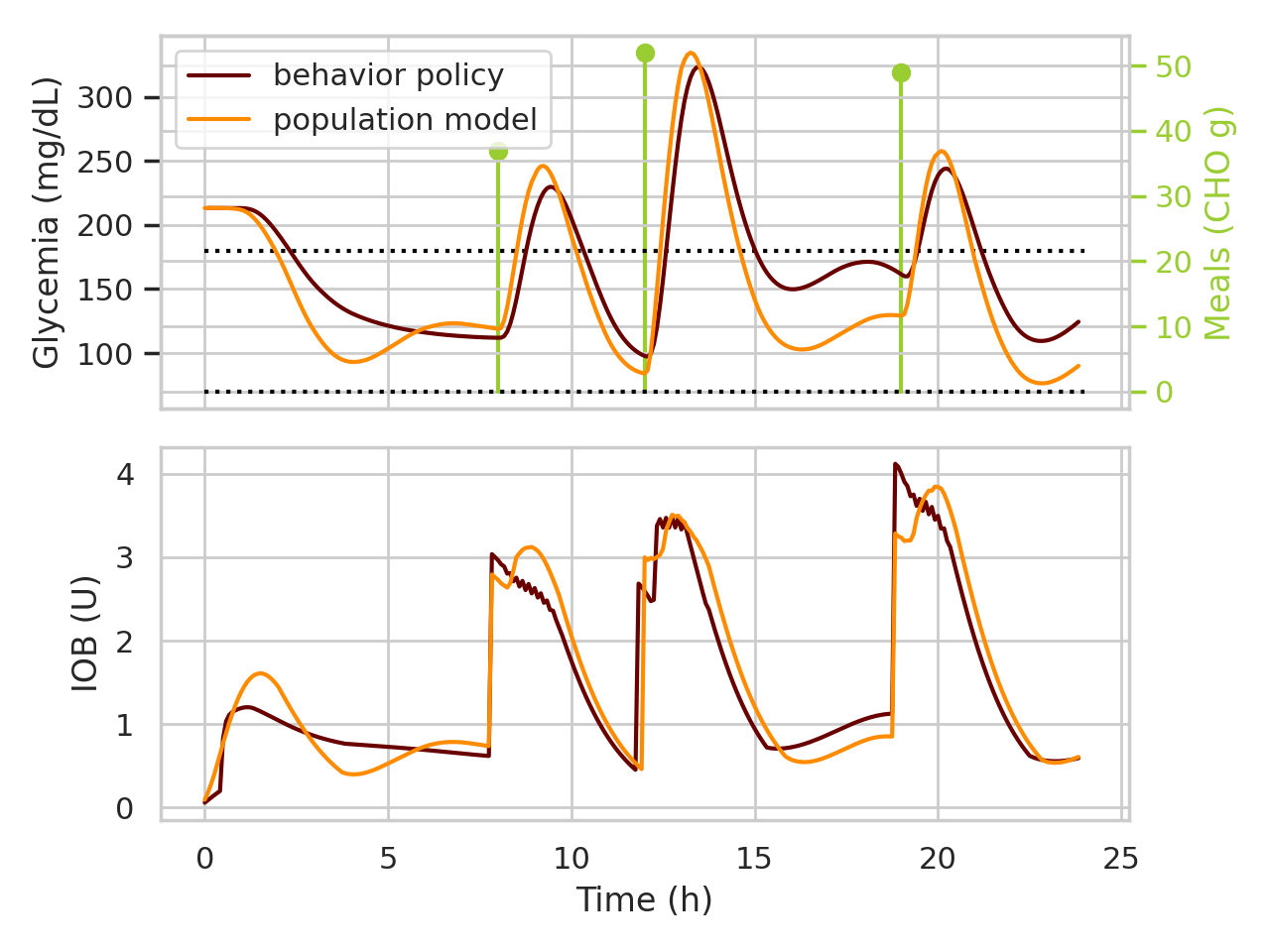}
  \caption{Comparison of the RL agent and the behavior policy on the simulator. The RL agent is more aggressive: at the start to reach a value close to 110 mg/dL, and also add more basal after the meal bolus. Mean glycemia is thus lower for the population model.}
  \label{fig:baseline_perf_sim}
\end{figure}

An example of control on the simulator is presented on Figure \ref{fig:baseline_perf_sim}. The observed differences are informative. The RL agent is much more aggressive in hyperglycemia situations, but also seems to anticipate more when the glycemia is high but decreasing. Also, during the meal periods, the RL agent is more aggressive, especially around 1h after the meal. 

The ability for a closed loop to deal with unannounced meals is of a paramount importance as it further decreases the burden of the disease on diabetic patients. To ensure that the population model is capable of handling such cases, we run a simulation with unannounced meals. In this scenario, the virtual patients don't declare any meals: the COB feature given to the offline agent and the behavior policy is always null and no meal bolus is automatically administered.

\begin{figure}[H]
  \centering
  \includegraphics[width=\textwidth]{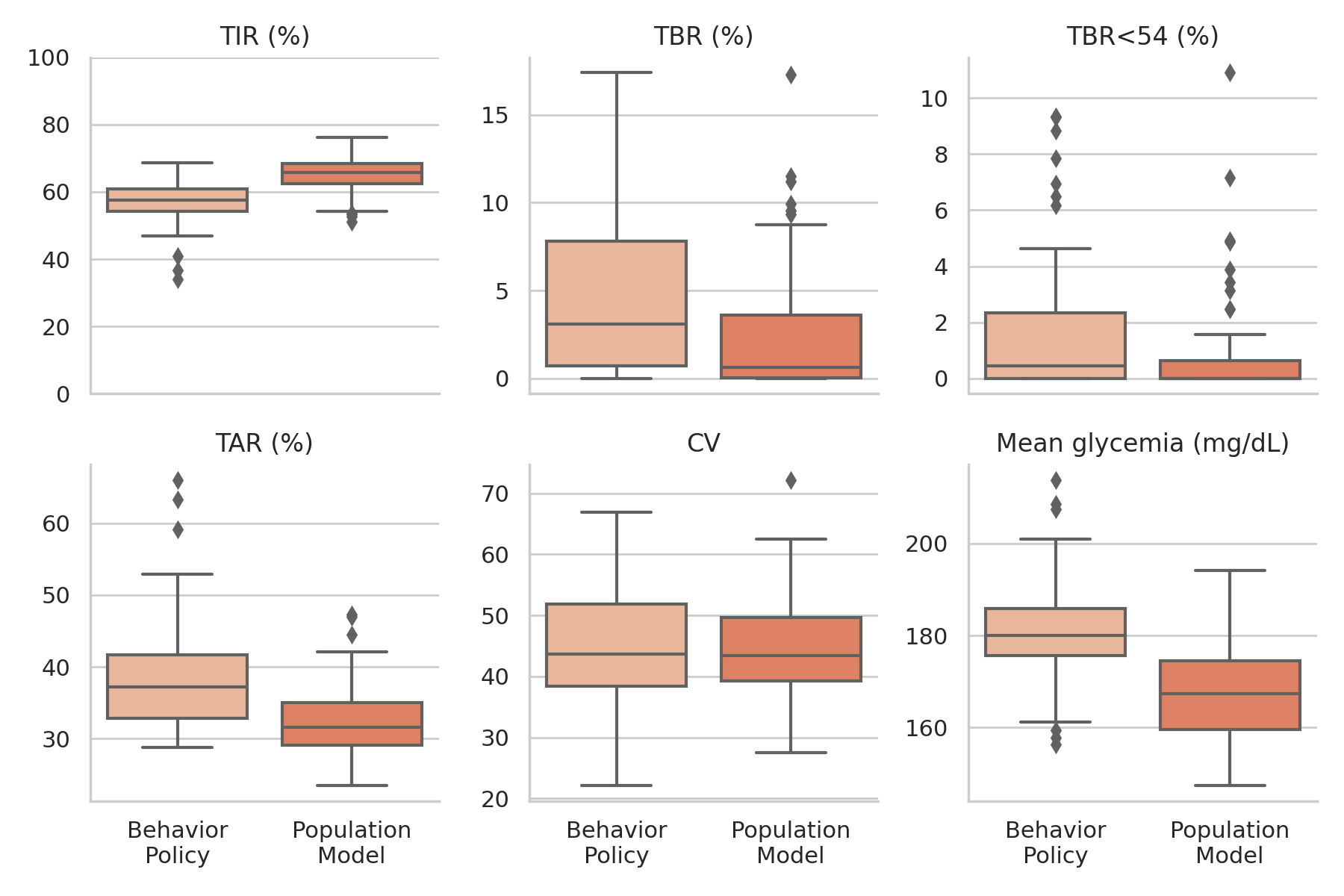}
  \caption{In silico comparison of behavior policy with personalized model in the case of unannounced meals. Simultaneous improvement of TIR, TBR, and mean glycemia.}
  \label{fig:dblg1_vs_dl_umeal}
\end{figure}

 Results presented on Figure \ref{fig:dblg1_vs_dl_umeal}, which compares the behavior policy and the population model. There is a significant improvement in the performance of our Reinforcement Learning (RL) agent compared to the behavior policy. Specifically, we observe a substantial 8.0\% increase in the TIR and a 6.1\% reduction in the mean TBR. Notably, the most significant improvement is observed in the mean glycemia, which decreases by an impressive 13.2 mg/dL. This reduction has promising implications for patient health, particularly in scenarii where meal announcements may be overlooked. These results mark an important initial step toward achieving a fully closed-loop glycemia control system, one that operates seamlessly without the need for meal declarations from patients.

\section{An end-to-end pipeline for patient-wise personalization}
\label{section:personalization}

In this section, we describe a pipeline for offline patient-wise model personalization. We start by proposing an application of fitted Q-evaluation which allows to estimate clinically relevant metrics. We then describe the experimental setup we propose and present the results.

\subsection{Fitted Q evaluation}

In order to evaluate an offline agent without the use of a simulator, off-policy evaluation methods can be used. Let $\pi$ the new policy of the agent to evaluate. Fitted Q evaluation (FQE) learns the Q function under the policy $\pi$ using Bellman equation \cite{le2019batch}:
$$Q(s_t, a_t) = c(s_t, a_t) + \gamma Q(s_{t+1}, \pi(s_{t+1}))$$
where $(s_t, a_t, s_{t+1})$ is a transition (state, action, next state) in the offline dataset and $c$ is a reward function. When training FQE to estimate the Q function under the new policy, the reward function $c$ can be different from the one used to train the agent. We leverage this possibility to devise a method which directly estimates TIR/TAR/TBR.

In particular, if the reward function $c = c_{\text{TIR}}$ is chosen as 
$$
c_{\text{TIR}}(x) = \begin{cases}
    1 & \text{if }  x\in [70,180] \text{ mg/dL} \\
    0 & \text{else } 
\end{cases}
$$

then 

$$\mathbb{E}_\pi[c_{\text{TIR}}(a,s)] = \text{TIR}_\pi$$
$$Q(s,a)  = \mathbb{E}\Bigg[\sum_{k=0}^{\infty} \gamma^k c_{\text{TIR}_{k}}(s_k,a_k)\Bigg] = \frac{\text{TIR}_\pi}{1-\gamma}$$
where $\text{TIR}_\pi$ is the TIR under the policy $\pi$, since the expected value of the reward is precisely the proportion of time spent in the normoglycemia range. Similarly, TBR and TAR can also be estimated with the following reward functions: $c_{\text{TBR}}(x) = \mathds{1}_{x<70}(x), c_{\text{TAR}}(x) = \mathds{1}_{x>180}(x)$. These choices of rewards for FQE allow to estimate clinically relevant metrics without any simulation. Note also that these estimations can be made in theory from any state, allowing for immediate short term predictions as well as global patient-wise estimations of TIR/TAR/TBR.

In \cite{le2019batch}, the authors showed that the estimated value obtained from the FQE algorithm is not always accurate but can effectively be used to rank different algorithms. This caveat is to keep in mind, and constitutes a limitation of this work. A detailed analysis of the performances of FQE depending on the distribution of data would be informative and will be part of further work.

\subsection{Personalization protocol}

A well-known limitation of current closed-loops is their high inter-patient variability, which can be seen on Figure \ref{fig:comparison_algos} and Table \ref{table:rl_comparison}.

\begin{figure}[H]
  \centering
  \includegraphics[width=\textwidth]{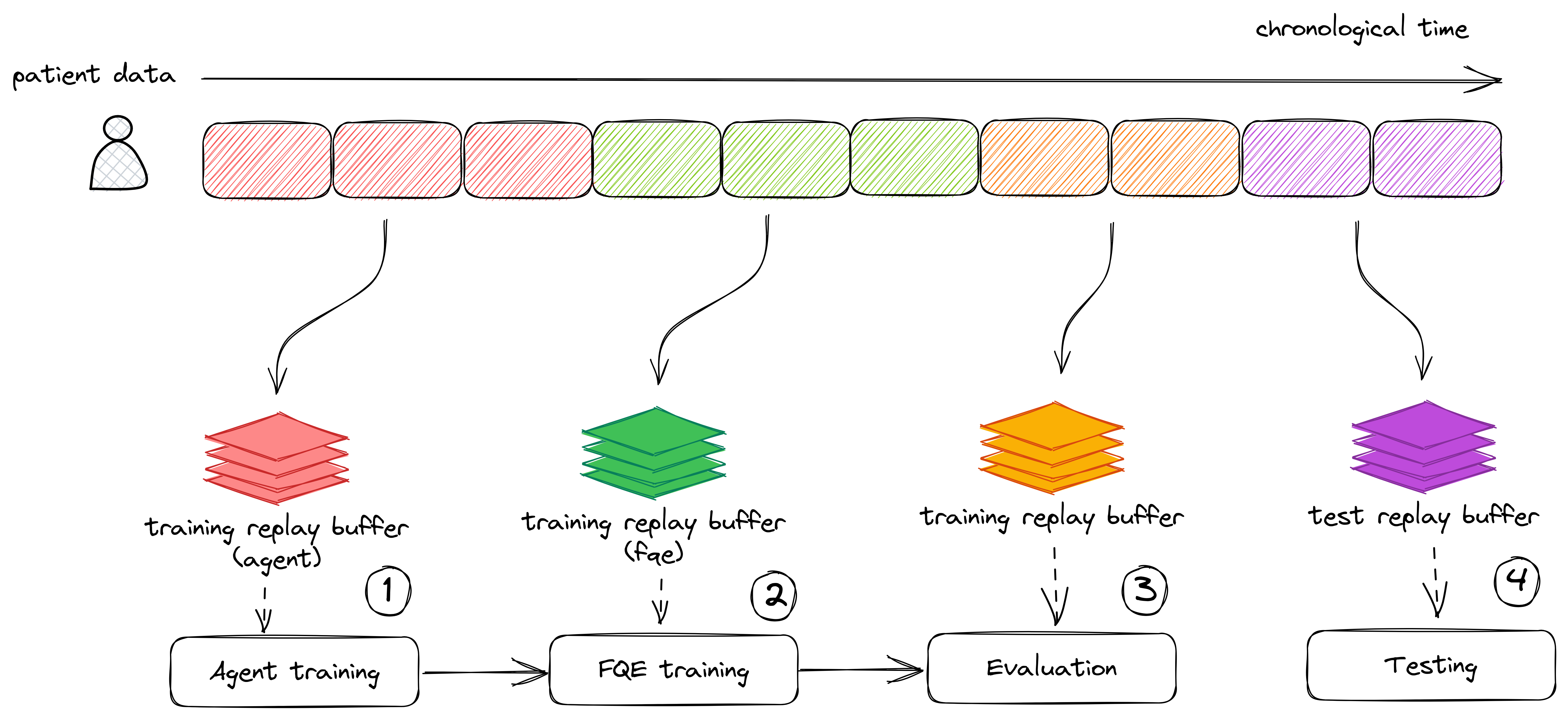}
  \caption{Personalization training procedure, each patient data is split into 4 chronological sets: one to train the agent on the patient data, one to train FQE model on, one to run validation during training and one to test the agent performance.}
  \label{fig:personalization_training_procedure}
\end{figure}

 To limit this issue, we propose a pipeline to personalize the agent to individual patient data. Starting from an offline RL agent, the training procedure is sketched on Figure \ref{fig:personalization_training_procedure}:
 \begin{enumerate}
    \item The first 25\% of the patient data is used as a data set for fine-tuning the agent,
    \item The following 25\% of the patient data is used to train a specific FQE model for each metric reward/TIR/TAR/TBR,
    \item The following 25\% of the patient data is used to evaluate the estimated metrics of the trained FQE models. The estimated reward serves for best checkpoint selection throughout the personalization, so as to prevent at best over-fitting and catastrophic forgetting,
    \item At the end of the training, the final 25\% of data is used to provide an unbiased measure of generalization performances, using the FQE models of the best checkpoint. 
 \end{enumerate}
 
%As FQE can only estimate q-values, it is not possible to evaluate offline the agent with the meal bolus calculator and the hypoglycemia prevention algorithms. This is the reason why performances from this section are lower than the previous one where online validation has been conducted -both meal bolus and hypoglycemia prevention were activated-.

In this series of experiments, we do not use the meal boluses and the hypoglycemia cutting logic: we evaluate the performances of the sole RL agent. This gives us a more accurate insight into the effect of the personalization of the whole control. This explains why the baseline performances in this section are lower (in fact, achieving such high TIR with an insulin rate lower than 10 U/h is a success). Of course, there is no obstacle in also running FQE evaluation with these features enabled.

To enable a fair comparison of the personalized models with the population model, we also train FQE models for reward/TIR/TBR/TAR for the population model, on the union of all data sets used for the FQE of the personalized models. We then evaluate these FQE models on the last 25\% data of each patient. These estimates of reward/TIR/TBR/TAR are the performances to beat.

\subsection{Results}

\paragraph{Overall results} We perform the personalization procedure on 25 patients separately. These patients are additional patients, not included in the training set of the population model. These patients have been observed on average 333 days, which means that about 3 months are used for the actual offline fine-tuning of the RL agent -the rest being used by the validation and test FQE for each metric. 

\begin{figure}[H]
  \centering
  \includegraphics[width=\textwidth]{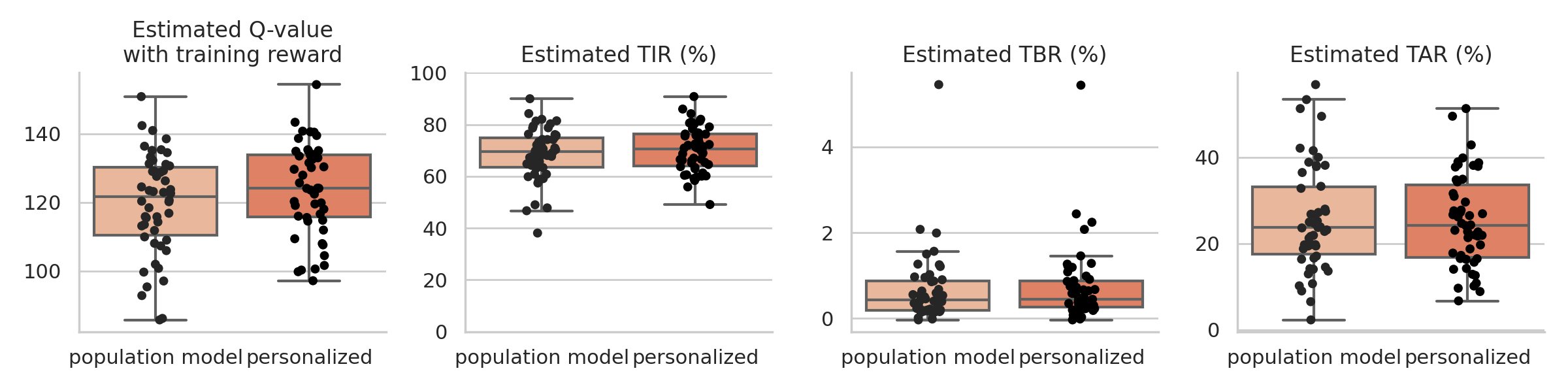}
  \caption{FQE estimation of the clinical metric before and after personalization, selecting the best model on training reward (Zhu).}
  \label{fig:perso_boxplot}
\end{figure}

We report the overall results on Figure \ref{fig:perso_boxplot}. All metrics improve on average with the personalization. The average Q-value estimated for the training reward increases by 10, the average TIR increases of 1\%, the average TBR increases by 1 \% and the TAR average is unchanged. Notably, worst cases are greatly improved e.g. the worst TIR is estimated to increase from 38 \% up to 50\%.

\begin{figure}[H]
  \centering
  \includegraphics[width=\textwidth]{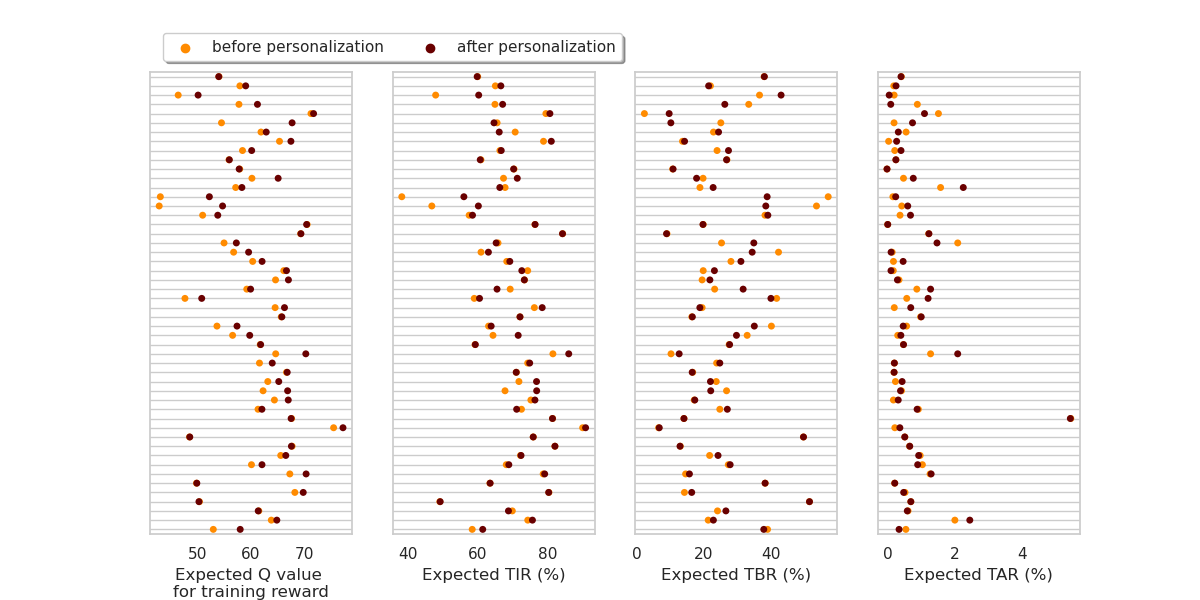}
  \caption{FQE estimation of each clinical metric before and after personalization for each patient. Almost all metrics improve for all patients.}
  \label{fig:perso_lollipop}
\end{figure}

Figure \ref{fig:perso_lollipop} gives the variations of each estimated metric for each patient between the population model and the personalized model. Note that although our checkpoint selection criterion throughout the personalization is based on the training reward, almost metrics do improve for almost all patients.

This shows that 3 months of data is enough to perform with success an off\-line patient-wise personalization of RL agents. Investigating more precisely the connection between the agents improvements and the quantity of data available would be another interesting continuation of this work. 

%\begin{figure}[H]
%  \centering
%  \includegraphics[width=0.3\textwidth]{images/personalization/sum_fqe.png}
%  \caption{Sum TIR, TAR, TBR, should sum up to 100\%}
%  \label{fig:perso_boxplot}
%\end{figure}

% \begin{figure}[H]
%   \centering
%   \includegraphics[width=\textwidth]{images/personalization_boxplot_best_on_tir.png}
%   \caption{FQE estimation of the clinical metric before and after personalization (the best model is selected with the TIR criteria instead of the training reward). TIR has significantly improved, pvalue of 0.02 for the test "the distribution of the TIR is higher for the best model than the baseline"}
%   \label{fig:perso_boxplot}
% \end{figure}

\begin{figure}[H]
  \centering
  \includegraphics[width=\textwidth]{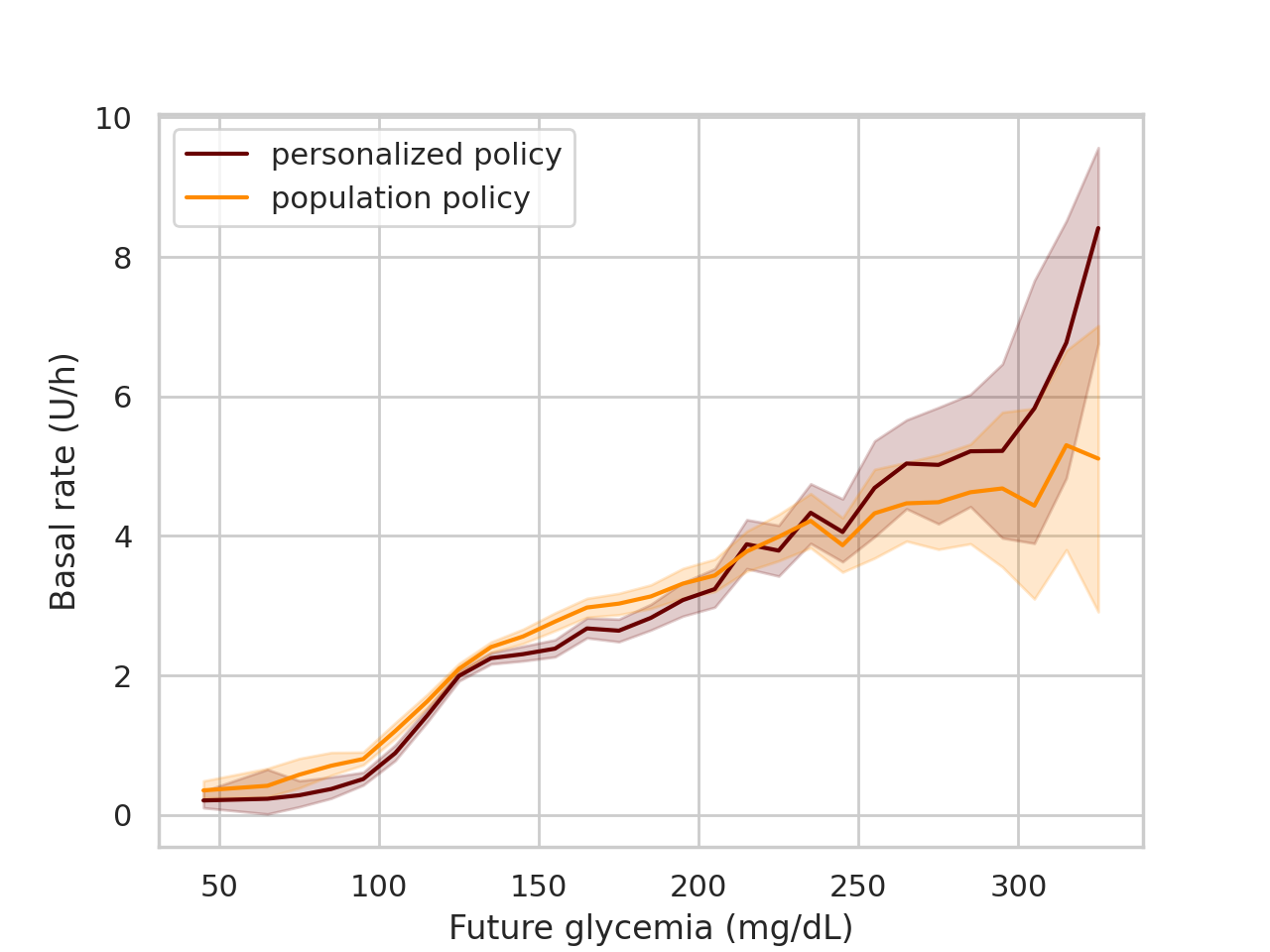}
  \caption{Variation of the prescribed basal rate between the population and personalized policy. X axis is the glycemia 30 minutes in the future. If the glycemia at t + 30 min is high (resp. low), the personalization is successful if it increased (resp. decreased) the basal rate with respect to the population model. This is what we observe here: the personalization led to increased basal rates above 200 mg/dL, and decreased basal rates below.}
  \label{fig:perso_analysis}
\end{figure}

\paragraph{Personalization analysis} To further evaluate how the personalization works, we use the fact that insulin ingested at time t has its largest impact on glycemia at time around t + 30 minutes. Therefore, if for some state in the real data, the glycemia 30 minutes later was high (resp. low), then the new control improved over the existing control if it suggested to send an higher (resp. lower) quantity of insulin 30 minutes before. We make this analysis on Figure \ref{fig:perso_analysis} which shows the basal rates delivered by the personalized models and the population model, with respect to the glycemia 30 minutes in the future. When the future glycemia is below 200 mg/dL, the personalized agents send less insulin on average. When the future glycemia is above 200 mg/dL, the personalized agents send more insulin on average. This constitutes another validation of the quality of the personalization. 

%\begin{figure}[H]
%  \begin{subfigure}[b]{0.5\textwidth}
%    \centering
%    \includegraphics[width=\linewidth]{images/personalization/improvement_while_hyper_id_36.png}
%    \caption{Increase of the basal rate for state with high glycemia (> 330 mg/dL). FQE predicts a decrease of 15\% of the TAR after personalization for this patient. This figure shows that the personalized policy should reduce the TAR as FQE predicted.}
%    \label{fig:perso_improvement_hyper}
%  \end{subfigure}
%  \begin{subfigure}[b]{0.5\textwidth}
%    \centering
%    \includegraphics[width=\linewidth]{images/personalization/improvement_while_hypo_id_36.png}
%    \caption{Steady or a bit less insulin sent during near hypoglycemic events (between 70 and 80 mg/dL)}
%    \label{fig:custom_reward}
%  \end{subfigure}
%  \caption{Improvement of the policy with personalization: the personalized policy sends more insulin than the population policy during hyper glycemia events \ref{fig:perso_improvement_hyper}, and not more when in hypoglycemia \ref{fig:perso_improvement_hypo}. As this patient has an estimated TAR $> 50\%$, this shows that the personalization procedure enabled to adapt to the patient physiology and reduce hyperglycemia without increasing TBR.}
%  \label{fig:perso_improvement_hypo}
%\end{figure}

\section{Conclusion}
\label{section:conclusion}

We performed an extensive comparison of offline RL algorithms for glycemia control on real data. The offline RL often outperforms the behavior policy. For example, the best TD3-BC model we trained has +7\% TIR, -1\% TBR and -12 mg/dL of mean glycemia on average, across a family of in silico patients. Further in silico evaluations with no announced meals illustrate the high robustness and accuracy of the RL control.

Going further, we showed how personalization of such RL agents can be made on individual patients in a realistic setting. This is illustrated using OPE methods, in a way that enables to recover key diabetics metrics directly -instead of a hard-to-interpret Q-value estimate. A couple of months of observations allows to improve the patients' TIR by an estimated 1 \% on average and to drastically reduce the variance of inter-subject performances. Such an improvement of worst cases of glycemia control in closed loop systems is of particular interest, as it is often a limitation of current commercial articifical pancreas.

Continuation of this work could include an ablation study to see if manual patient actions within the dataset do improve to the offline RL trainings. Additionally, a more rigorous evaluation of the FQE method applied to TIR/TBR/TAR estimation may be interesting.

\section*{Acknowledgements}

The completion of this paper has been made possible thanks to the support of Diabeloop SA. The data used in the paper has been provided by Diabeloop SA while complying with the RGPD reglementation. We also thank Pierre Gauthier for his contributions on the simulator-related experiments in the paper.

\newpage

\bibliographystyle{elsarticle-num} %unsrt
\bibliography{biblio}

% Ideas
% Give the experiments to design the reward function in the appendix
% Experiments on real data for reward choice also
% Test with ablation of all manual boluses and open loop transitions
% Show different meal patterns and correlate them to how the control is made
% e.g. do some patients under announce ?
% Illustrate the effectiveness of recovering tir/hypo/hyper.
% age et comportement de personalization
% 

\appendix

\end{document}